\pdfoutput=1

\documentclass[11pt]{article}

\usepackage{emnlp2021}

\usepackage{times}
\usepackage{latexsym}
\usepackage{graphicx}
\usepackage[T1]{fontenc}

\usepackage[utf8]{inputenc}

\usepackage{microtype}
\usepackage{comment}
\newcommand{\stitle}[1]{\vspace{1ex}\noindent{\bf #1}}

\usepackage{xspace}
\usepackage{xcolor}
\usepackage{mathtools}
\usepackage{bm}
\usepackage{multirow}
\usepackage{booktabs}
\usepackage{esvect}
\usepackage{amsfonts}
\usepackage{cleveref}
\crefformat{section}{\S#2#1#3}
\crefformat{subsection}{\S#2#1#3}
\crefformat{subsubsection}{\S#2#1#3}
\crefrangeformat{section}{\S\S#3#1#4 to~#5#2#6}
\crefmultiformat{section}{\S\S#2#1#3}{ and~#2#1#3}{, #2#1#3}{ and~#2#1#3}
\usepackage{refstyle}
\Crefformat{figure}{#2Fig.~#1#3}
\Crefmultiformat{figure}{Figs.~#2#1#3}{ and~#2#1#3}{, #2#1#3}{ and~#2#1#3}
\Crefformat{table}{#2Tab.~#1#3}
\Crefmultiformat{table}{Tabs.~#2#1#3}{ and~#2#1#3}{, #2#1#3}{ and~#2#1#3}
\Crefformat{appendix}{\S#2#1#3}
\crefformat{algorithm}{Alg.~#2#1#3}
\Crefformat{equation}{Eq.~#2#1#3}

\newcommand{\seg}{\textsc{EventSeg}\xspace}

\DeclareMathOperator{\R}{ReLU}

\title{
\mbox{Learning Constraints and Descriptive Segmentation for Subevent Detection}}

\author{Haoyu Wang$^1$, Hongming Zhang$^2$\thanks{\indent This work was done when the author was visiting the University of Pennsylvania.}, Muhao Chen$^{3}$ \& Dan Roth$^1$\\
$^1$Department of Computer and Information Science, UPenn\\
$^2$Department of Computer Science and Engineering, HKUST\\
$^3$Department of Computer Science, USC\\
\texttt{\{why16gzl,danroth\}@seas.upenn.edu};\\ \texttt{hzhangal@cse.ust.hk};
\texttt{muhaoche@usc.edu}\\
}
\begin{document}
\maketitle
\begin{abstract}

Event mentions in text correspond to real-world events of varying degrees of granularity.
The task of subevent detection aims to resolve this granularity issue, recognizing the membership of multi-granular events in event complexes. 
Since knowing the span of descriptive contexts of event complexes helps infer the membership of events, we propose the task of \emph{event-based text segmentation} (\seg) as an auxiliary task to improve the learning for subevent detection.
To bridge the two tasks together, we propose an approach to learning and enforcing constraints that capture dependencies between subevent detection and \seg prediction, as well as guiding the model to make globally consistent inference.
Specifically, we adopt Rectifier Networks for constraint learning and then convert the learned constraints to a regularization term in the loss function of the neural model. 
Experimental results show that the proposed method outperforms baseline methods by 2.3\% and 2.5\% on benchmark datasets for subevent detection, HiEve and IC, respectively, while achieving a decent performance on \seg prediction\footnote{Our code is publicly available at \url{http://cogcomp.org/page/publication_view/950}.}. 
\end{abstract}

\section{Introduction}


Since real-world events are frequently conveyed in human languages, understanding their linguistic counterparts, i.e. event mentions in text, is of vital importance to natural language understanding (NLU).
One key challenge to understanding event mentions is that they refer to real-world events with varied granularity \cite{glavas-etal-2014-hieve} and form \emph{event complexes} \cite{wang-etal-2020-joint}.
For example, when speaking of a coarse-grained event ``publishing a paper'', it can involve a complex of more fine-grained events such as ``writing the paper,'' ``passing the peer review,'' and ``presenting at the conference.''
Naturally, understanding events requires to resolve the granularity of events and infer their memberships, which corresponds to the task of subevent detection (a.k.a. event hierarchy extraction).
Practically, subevent detection is a key component of event-centric NLU \cite{chen2021event-acl}, and is beneficial to various applications, such as 
schema induction \cite{zhang-etal-2020-analogous,li-etal-2020-connecting}, task-oriented dialogue agents \cite{andreas2020task}, summarization \cite{ijcai2019-0686, zhao-etal-2020-improving}, and risk detection \cite{pohl2012automatic}.

\begin{figure}[t]
 \centering
 \includegraphics[scale=0.5]{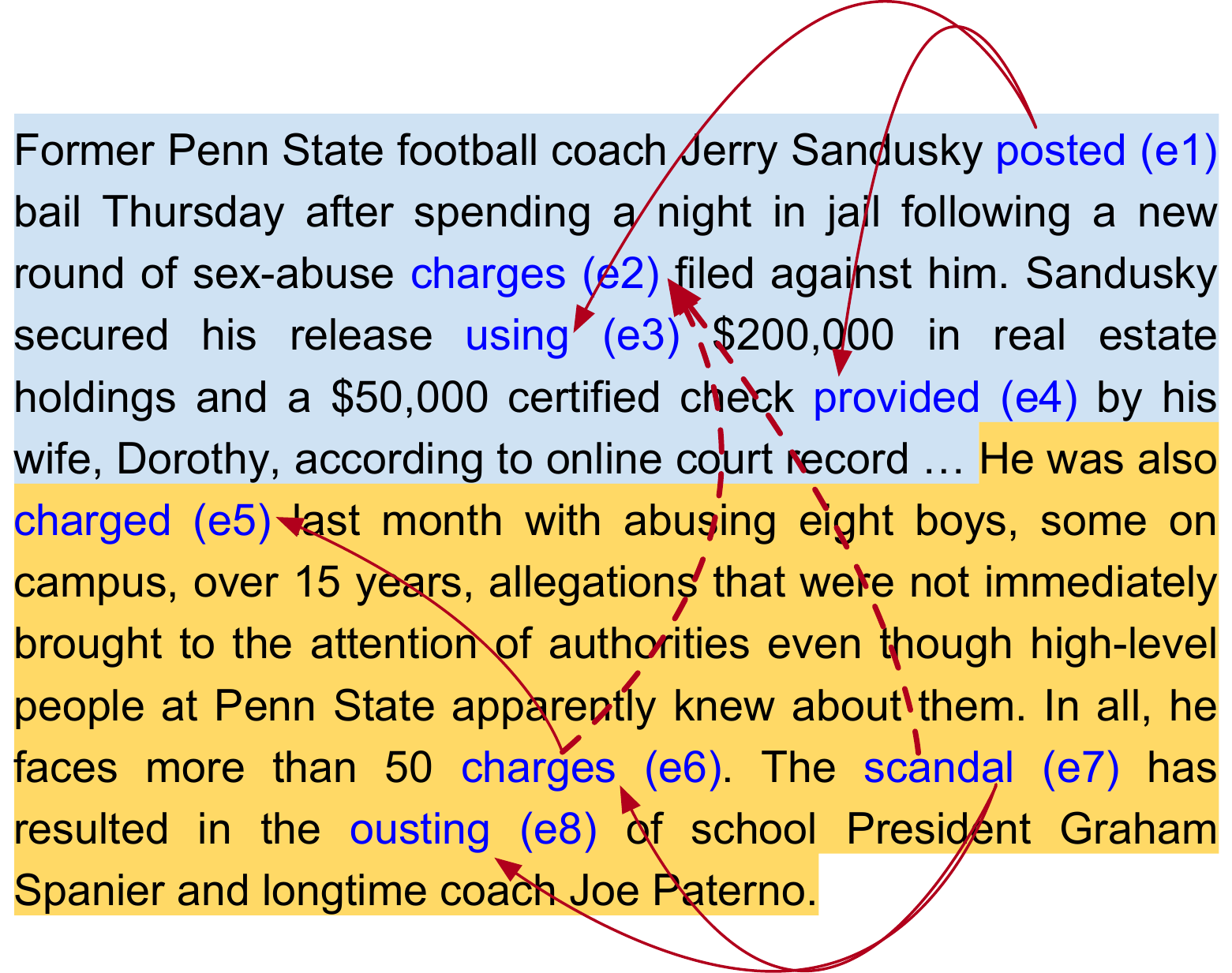}
 \caption{An example of \textsc{Parent-Child} relations and \textsc{EventSeg}s from the HiEve dataset \cite{glavas-etal-2014-hieve}. The blue and yellow segments denote the textual spans of event complexes ``posted'' and ``scandal'' respectively. Curved arrows denote \textsc{Parent-Child} relations within a text segment, whereas the dotted arrows denote cross-segment \textsc{Parent-Child} relations.}
 \label{fig:example}
\end{figure}

As a significant step towards inducing event complexes (graphs that recognize the relationship of multi-granular events) in documents, subevent detection has started to receive attention recently \cite{wang-etal-2020-joint, han2021ester}. 
It is natural to perceive that in documents, there might be several different event complexes and they often span in different descriptive contexts that form relatively independent text segments.
Consider the example in \Cref{fig:example}, where the two membership relations in the event complex (graph consisting of ``scandal (e7),'' ``charges (e6),'' ``ousting (e8),'' and relations) are both within the segment marked in yellow that describes the event complex.
As can be seen in the paragraph, though we cannot deny the existence of cross-segment subevent relations (dotted arrows), 
events belonging to the same membership are much more often to co-occur in a text segment.
This correlation has been overlooked by existing data-driven methods \cite{zhou-etal-2020-temporal,yao-etal-2020-weakly}, which formulate subevent detection as pairwise relation extraction.
On the other hand, while prior studies have demonstrated the benefits of incorporating logical constraints among event memberships and other relations (such as co-reference) \cite{glavas-snajder-2014-constructing, wang-etal-2020-joint}, the constraints between the memberships and event co-occurences in text segments remain uncertain.
Hence, how to effectively learn and enforce hard-to-articulate constraints as in the case of subevent detection and segmentation of text is another challenge.


Our \emph{first} contribution is to improve subevent detection based on an auxiliary task of \seg prediction.
By \seg prediction, we seek to segment a document into descriptive contexts of different event complexes.
Evidently, with \seg information, it would be relatively easy to infer the memberships of events in the same descriptive context.
Using annotations for subevent detection and \seg prediction, we aim to adopt a neural model to jointly learn these two tasks along with the (soft) logical constraints that bridge their labels together.
In this way, we incorporate linear discourse structure of segments into membership relation extraction, avoiding complicated feature engineering in the previous work \cite{aldawsari-finlayson-2019-detecting}.
From the learning perspective, adding \seg prediction as an auxiliary task seeks to provide effective incidental supervision signals \cite{Roth17} to the subevent detection task. This is especially important in the current scenario where annotated learning resources for subevents are typically limited \cite{hovy-etal-2013-events, glavas-etal-2014-hieve, ogorman-etal-2016-richer}.


To capture the logical dependency between subevent structure and \seg,
our \emph{second} contribution is an approach to automatically learning and enforcing logical constraints. 
Motivated by \citet{pan-etal-2020-learning}, we use Rectifier Networks to learn constraints in the form of linear inequalities, and then convert the constraints to a regularization term that can be incorporated into the loss function of the neural model.
This allows any hard-to-articulate constraints to be automatically captured for interrelated tasks, and efficiently guides the model to make globally consistent inference.
By learning and enforcing task-specific constraints for subevent relations, the proposed method achieves comparable results with SOTA subevent detection methods on the HiEve and IC dataset.
Moreover, by jointly learning with \seg prediction, the proposed method surpasses previous methods on subevent detection by relatively 2.3\% and 2.5\% in $F_1$ on HiEve and IC, while achieving decent results on \seg prediction.

\section{Related Work}

We discuss three lines of relevent research.

\stitle{Subevent Detection.} Several approaches to extracting membership relations have been proposed, which mainly fall into two categories: statistical learning methods and data-driven methods.
Statistical learning methods \cite{glavas-etal-2014-hieve, glavas-snajder-2014-constructing, araki-etal-2014-detecting, aldawsari-finlayson-2019-detecting} collect a variety of features before feeding into classifiers for pairwise decision. 
Nevertheless the features 
often require costly human effort to obtain,
and are often dataset-specific. 
Data-driven methods, on the other hand, automatically characterize events with neural language models like BERT \cite{devlin-etal-2019-bert}, and can simultanously incorporate various signals such as event time duration \cite{zhou-etal-2020-temporal}, joint constraints with event temporal relations \cite{wang-etal-2020-joint} and subevent knowledge \cite{yao-etal-2020-weakly}.
Among recent methods, only \citet{aldawsari-finlayson-2019-detecting} utilize discourse features like discourse relations between elementary discourse units, but still document-level segmentation signals are not incorporated into the task of subevent detection.
Actually, 
research on event-centric NLU \cite{chen2021event-acl} has witnessed the usage of document-level discourse relations: different functional discourse structures around the main event in news articles have been studied in \citet{choubey-etal-2020-discourse}.
Hence, we attempt to capture the interdependencies between subevent detection and segmentation of text, in order to enhance the model performance for event hierarchy extraction.

\stitle{Text Segmentation.}
Early studies in this line have concentrated on unsupervised text segmentation, quantifying lexical cohesion within small text segments \cite{choi-2000-advances}, and unsupervised Bayesian approaches have also been successful in this task \cite{eisenstein-barzilay-2008-bayesian, eisenstein-2009-hierarchical, newman-etal-2012-bayesian, mota-etal-2019-beamseg}.
Given that unsupervised algorithms are difficult to specialize for a particular domain, \citet{koshorek-etal-2018-text} formulate the problem as a supervised learning task.
\citet{lukasik-etal-2020-text} follow this idea by using transformer-based architectures with cross segment attention to achieve state-of-the-art performance.
Focusing on creating logically coherent sub-document units, these prior work do not cover segmentation of text regarding descriptive contexts of event complexes, which is the focus of the auxiliary task in this work. 

\stitle{Learning with Constraints.}
In terms of enforcing declarative constraints in neural models, early efforts \cite{RothYi04, glavas-snajder-2014-constructing} formulate the inference process as Integer Linear Programming (ILP) problems. 
\citet{pan-etal-2020-learning} also employ ILP to enforce constraints learned automatically from Rectifier Networks with strong expressiveness \cite{pmlr-v48-panb16}.
Yet the main drawback of solving an ILP problem is its inefficiency in a large feasible solution space.
Recent work on integrating neural networks with structured outputs has emphasized the importance of the interaction between constraints and representations \cite{NIPS2017_b2ab0019, pmlr-v80-niculae18a, li-srikumar-2019-augmenting, li-etal-2019-logic, li-etal-2020-structured}.
However there has been no automatic and efficient ways to learn and enforce constraints that are not limited to first-order logic, e.g., linear inequalities learned via Rectifier Networks.
And this is the research focus of our paper.

\section{Preliminaries}\label{sec:prelim}

A document $\mathcal{D}$ consists of a collection of $m$ sentences $\mathcal{D} = [s_1, s_2, \cdots, s_m]$, and each sentence, say $s_k$, contains a sequence of tokens $s_k = [w_1, w_2, \cdots, w_n]$. 
Some tokens in sentences belong to the set of annotated event triggers, i.e., $\mathcal{E}_{\mathcal{D}} = \{e_1, e_2, \cdots, e_l\}$. 
Following the notation by \citet{koshorek-etal-2018-text}, a segmentation of document $\mathcal{D}$ is represented as a sequence of binary values: $\mathcal{Q}_{\mathcal{D}} = \{q_1, q_2, \cdots, q_{m-1}\}$, where $q_i$ indicates whether sentence $s_i$ is the end of a segment.

\stitle{Subevent Detection} is to 
identify membership relations between events, given event mentions in documents.
Particularly, $\mathcal{R}$ denotes the set of relation labels as defined in \citet{hovy-etal-2013-events} and \citet{glavas-etal-2014-hieve} (i.e., \textsc{Parent-Child, Child-Parent}, \textsc{Coref}, and \textsc{NoRel}). For a relation $r \in \mathcal{R}$, we use a binary indicator $Y_{i, j}^r$ to denote whether an event pair $(e_i, e_j)$ has relation $r$, and use $y_{i, j}^r$ to denote the model-predicted possibility of an event pair $(e_i, e_j)$ to have relation $r$. 

\stitle{EventSeg} prediction aims at finding an optimal segmentation of text that breaks the document into several groups of consecutive sentences, and each sequence is a descriptive context of an event complex \cite{wang-etal-2020-joint}. 
Being different from the traditional definition of text segmentation, \seg focuses on the change of event complex (which is not necessarily the change of topic).
For a pair of events $(e_i, e_j)$, we use a binary indicator $Z_{i, j}$ to denote whether the two events are within the same descriptive context of event complexes, and $z_{i, j}$ to denote the model-predicted possibility of two events to belong to the same segment.
Details on how to obtain \seg are described in \Cref{sec:obtain_seg}.

\stitle{Connections between Two Tasks.} 
Statistically, through an analysis of the HiEve and IC corpus, \textsc{Parent-Child} and \textsc{Child-Parent} relations appear within the same descriptive context of event complex with a probability of 65.13\% (see \Cref{tab:stats}). 
On the other hand, the probability for each of the two other non-membership relations (i.e., \textsc{Coref} and \textsc{NoRel}) to appear within the same segment approximately equals that of 
its appearence across segments.
This demonstrates that subevent relations tend to appear within the same \seg. 
Since this is not an absolute logical constraint, we adopt an automatic way of modeling such constraints instead of manually inducing them, which is described in the next section.
\section{Methods}

We now present the framework for learning and enforcing constraints for the main task of subevent detection and the auxiliary \seg prediction. 
We start with learning the hard-to-articulate constraints (\Cref{sec:cons_learn}), followed by 
details of joint learning  (\Cref{sec:neural_model}) and inference (\Cref{sec:inference}) for the two tasks.

\subsection{Learning Constraints} \label{sec:cons_learn}

\begin{figure}
    \centering
    \includegraphics[scale=0.4]{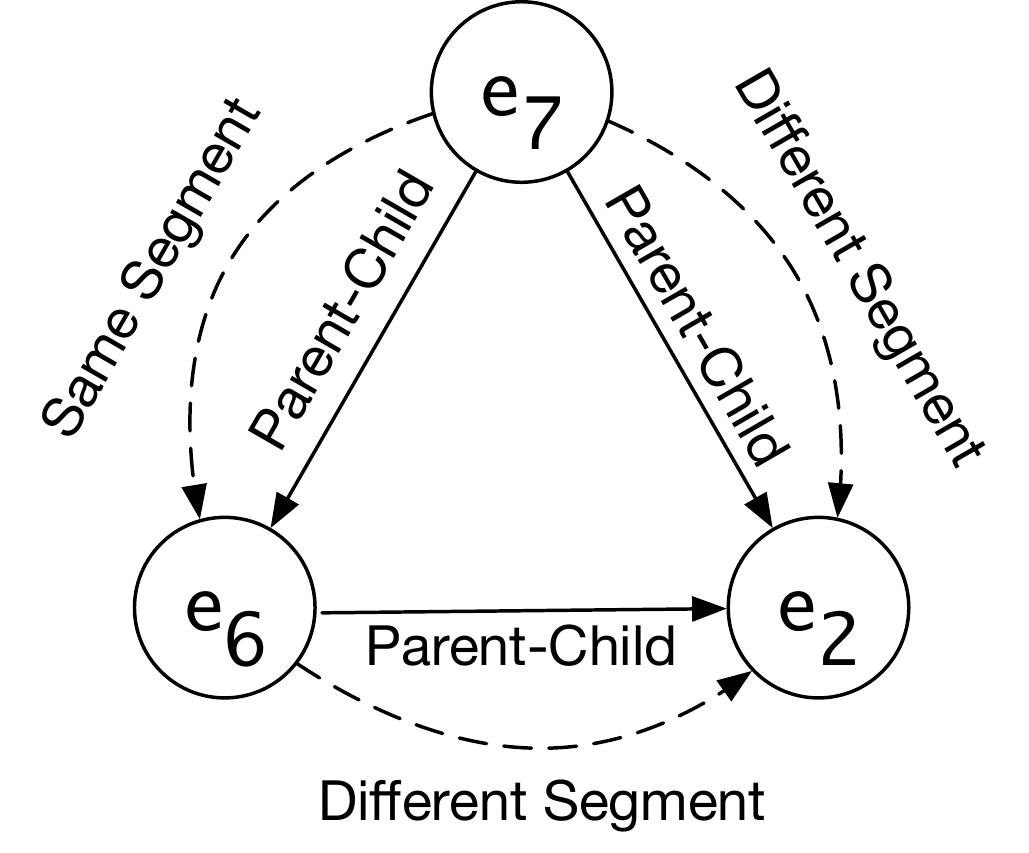}
    \caption{A legitimate structure for three-event subgraph obtained from the example shown in \Cref{fig:example}. The constraint features for the subgraph can be expressed by $\mathbf{X} = \mathbf{X}_{7, 6} \cup \mathbf{X}_{6, 2} \cup \mathcal{P}(\mathbf{X}_{7, 2})$, and the label $t$ for this structure is 1.}
    \label{fig:SDD}
\end{figure}

From the example shown in \Cref{fig:example} we can construct an event graph $G$ with all the events, membership relations, and \seg information. 
\Cref{fig:SDD} shows a three-event subgraph  of $G$. 
The goal of constraint learning is as follows: given membership relations $Y_{i, j}^r, Y_{j, k}^r$ and segmentation information $Z_{i, j}, Z_{j, k}$ about event pairs $(e_i, e_j)$ and $(e_j, e_k)$, we would like to determine whether a certain assignment of $Y_{i, k}^r, $ and $Z_{i, k}$ is legitimate.

\stitle{Feature Space for Constraints.}
We now define the feature space for constraint learning.
Let $\mathbf{X}_p = \{Y_p^r, r \in \mathcal{R}\} \cup \{Z_p\}$ denote the set of features for an event pair $p$. 
Given features $\mathbf{X}_{i, j}$ and $\mathbf{X}_{j, k}$, we would like to determine the value of $\mathbf{X}_{i, k}$,
yet the mapping from the labels of $(e_i, e_j), (e_j, e_k)$ to the labels of $(e_i, e_k)$ is a one-to-many relationship. 
For instance, if $r =$  \textsc{Parent-Child}, $Y_{i, j}^r = Y_{j, k}^r = 1$, and $Z_{i, j} = Z_{j, k} = 0$, then due to the transitivity of \textsc{Parent-Child}, we should enforce $Y_{i, k}^r = 1$. 
Yet we cannot tell whether $e_i$ and $e_k$ are in the same \seg, i.e., both $Z_{i, k} = 1$ and $Z_{i, k} = 0$ could be legitimate. 
In other words, we actually want to determine the \emph{set of possible values} of $\mathbf{X}_{i, k}$ and thus we need to expand the constraint features to better capture relationship legitimacy.
We employ the \emph{power set} of $\mathbf{X}_{i, k}$, $\mathcal{P}(\mathbf{X}_{i, k})$, as our new features for event pair $(e_i, e_k)$. 
And now a subgraph with three events $e_i$, $e_j$, and $e_k$ can be featurized as
\begin{equation}
    \mathbf{X} = \mathbf{X}_{i, j} \cup \mathbf{X}_{j, k} \cup \mathcal{P}(\mathbf{X}_{i, k}).
\end{equation}

\stitle{Constraint Learning with Rectifier Network.}
When we construct three-event subgraphs from documents, a binary label $t$ for structure legitimacy is created for each subgraph.
Inspired by how constraints are learned for several structured prediction tasks  \cite{pan-etal-2020-learning}, we represent constraints for a given subgraph-label pair $(\mathbf{X}, t)$ as $K$ linear inequalities.\footnote{Here we assume $K$ constraints is the upper bound for all the rules to be learned.}
Formally, $t=1$ if $\mathbf{X}$ satisfies constraints $c_k$ for all $k=1,\cdots,K$.
And the $k^\text{th}$ constraint $c_k$ is expressed by a linear inequality
\begin{align*}
    \begin{split}
        \mathbf{w}_k \cdot \mathbf{X} + b_k \ge 0,
    \end{split}
\end{align*}
whose weights $\mathbf{w}_k$ and bias $b_k$ are learned.
Since a system of linear inequalities is proved to be equivalent to the Rectifier Network proposed in \citet{pan-etal-2020-learning}, we adopt a two-layer rectifier network for learning constraints
\begin{equation}\label{relu_network}
p = \sigma\Big(1- \sum_{k=1}^K \R \big(\mathbf{w}_k \cdot \mathbf{X} + b_k\big)\Big),
\end{equation}
where $p$ denotes the possibility of $t = 1$ and $\sigma(\cdot)$ denotes the sigmoid function.
We train the parameters $\mathbf{w}_k$'s and $b_k$'s of the rectifier network in a supervised setting. The positive examples are induced from subgraph structures that appear in the training corpus, while the negative examples are randomly chosen from the rest possibilities that do not exist in the training corpus. 

\subsection{Joint Task Learning} \label{sec:neural_model}
\begin{figure*}
    \centering
    \includegraphics[width=\linewidth]{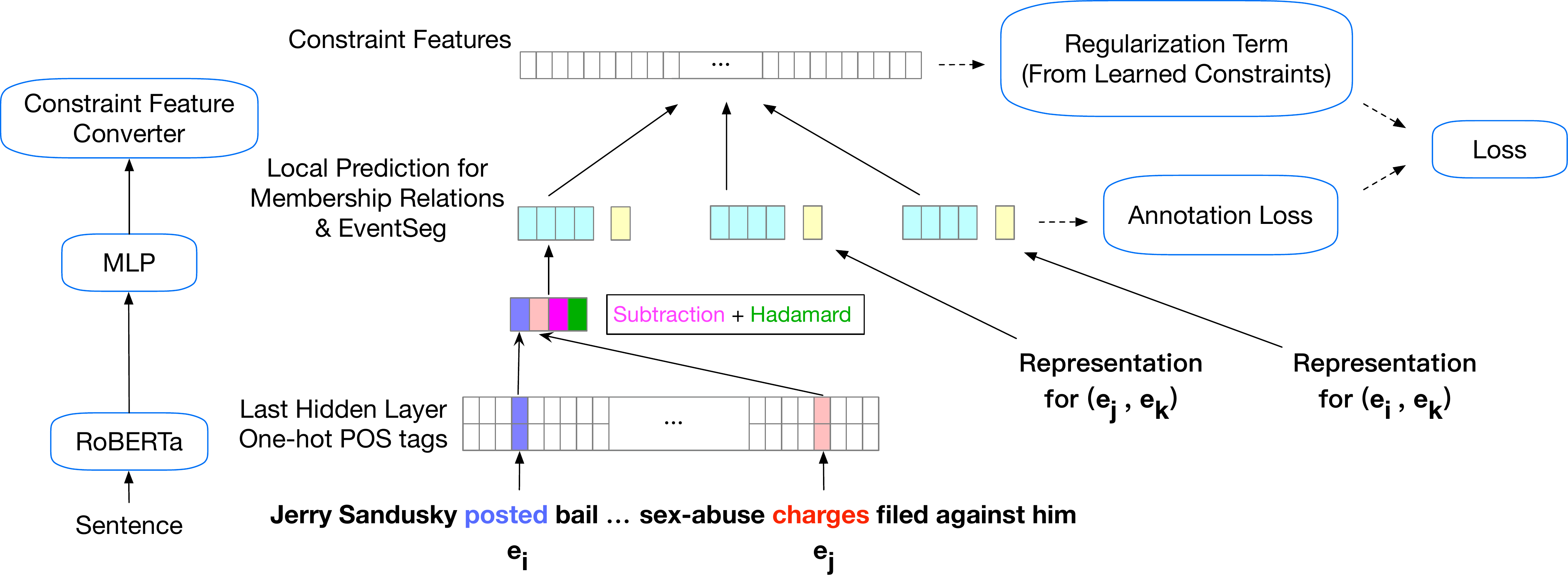}
    \caption{An overview of our approach. The model takes three pairs of events at a time in training to enforce constraints over three-event subgraphs (an example can be found in \Cref{fig:SDD}). Event pair representations are obtained from RoBERTa where the context of two events are taken into consideration. Soft logical constraints learned in \Cref{sec:cons_learn} are converted to a regularization term in the loss function for  subgraph structure legitimacy.}
    \label{fig:model}
\end{figure*}

After learning the constraints using Rectifier Networks, we introduce how to jointly model membership relations and \seg with neural networks and how to integrate the learned constraints into the model.
The model architecture is shown in \Cref{fig:model}.

\stitle{Local Classifier.}
To characterize event pairs in documents, we employ a neural encoder, which obtains contextualized representations for event triggers from the pre-trained transformer-based language model RoBERTa \cite{liu2019roberta}. 
As the context of event pairs, the sentences where two event mentions appear are concatenated using [CLS] and [SEP].
We then calculate the element-wise average of subword-level contextual representations as the representation for each event trigger.
To obtain event pair representation for $(e_i, e_j)$, we concatenate the two contextual representations, together with their element-wise Hadamard product and subtraction as in \citet{wang-etal-2020-joint}.
The event pair representation is then sent to a multi-layer perceptron (MLP) with $|\mathcal{R}|$ outputs for estimation of the confidence score $y_{i, j}^r$ for each relation $r$.
To make \seg as an auxiliary task, the model also predicts whether two events belong to the same segment using another separate MLP with a single-value output $z_{i, j}$.
In accordance with the learned constraints in \Cref{sec:cons_learn}, the model takes three pairs of events at a time.
The annotation loss in \Cref{fig:model} is a linear combination of a four-class cross-entropy loss $L_{A, sub}$ for subevent detection and a binary cross-entropy loss $L_{A, seg}$ for \seg.

\stitle{Incorporating Subgraph Constraints.} 
The $K$ constraints learned in \Cref{sec:cons_learn} are encoded into the weights $\mathbf{w}_k$ and bias $b_k$, $k = 1, \cdots, K$. 
Now that the input $\mathbf{X}$ is considered valid if it satisfies all $K$ constraints, we obtain the predicted probability $p$ of $\mathbf{X}$ being valid from \Cref{relu_network}.
To add the constraints as a regularization term in the loss function of the neural model, we convert $p$ into the negative log space \cite{li-etal-2019-logic} which is same as the cross-entropy loss.
And thus the loss corresponding to the learned constraints is
\begin{align*}\small
\begin{split}
L_{cons} = - log\Big(Sigmoid \big(1-\sum_{k=1}^N ReLU(\mathbf{w}_k \cdot \boldsymbol{\psi} + b_k)\big)\Big).
\end{split}
\end{align*}
And the loss function of the neural model is
\begin{equation}
    L = \lambda_1 L_{A, sub} + \lambda_2 L_{A, seg} + \lambda_3 L_{cons},
    \label{eq:loss}
\end{equation}
where the $\lambda$'s are non-negative coefficients to control the influence of each loss term.
With the loss function in \Cref{eq:loss}, 
we train the model in a supervised way to fine-tune RoBERTa.

\subsection{Inference}\label{sec:inference}
At inference time, to extract relations in the subevent detection task, we input a pair of events into the model and compare the predicted possibility for each relation, leaving the other two input pairs blank. 
For \seg prediction, we let the model predict $z_{i, i+1}$ for each pair of adjacent events $(e_i, e_{i+1})$ that appear in different sentences. 
If $z_{i, i+1} = 1$, it means there is a segment break between $e_i$ and $e_{i+1}$. When there are intermediate sentences between the two adjacent event mentions, we treat the sentence that contains $e_i$ as the end of a previous segment.
In this way, we provide an approach to solving two tasks together via 
automatically learning and enforcing constraints in the neural model. 
We provide in-depth experimentation for the proposed method in the next section.
\section{Experiments}

Here we describe the experiments on subevent detection with \seg prediction as an auxiliary task.
We first introduce the corpora used (\Cref{sec:datasets}), followed by evaluation for subevent detection and an ablation study for illustrating the importance of each model component (\Cref{sec:baselines}-\Cref{sec:results}).
We also provide a case study on \seg prediction (\Cref{sec:case_study}) and an analysis of the constraints learned in the model (\Cref{sec:exp_cons}).

\subsection{Datasets} \label{sec:datasets}

\begin{table}[]

\resizebox{\columnwidth}{!}{%
\begin{tabular}{c|cc|cc}
\hline
\toprule
\multirow{2}{*}{Relations} & \multicolumn{2}{c|}{HiEve} & \multicolumn{2}{c}{IC} \\ 
                  & Within       & Across       & Within      & Across     \\ \hline
Parent-Child      & 1,123         & 679         & 1,698        & 550       \\ 
Child-Parent      & 1,067         & 779         & 1,475        & 863       \\ 
Coref             & 322          & 436         & 1,476        & 877       \\ 
NoRel             & 32,029        & 31,726       & 40,072       & 41,815     \\ 
\bottomrule
\end{tabular}
}
\caption{Statistics of the HiEve and IC dataset. Numbers in column ``Within'' denote the number of relations appearing within the same descriptive context of event complex, whereas numbers under ``Across'' denote those across different segments.}
\label{tab:stats}
\end{table}
\stitle{HiEve}
The HiEve corpus \cite{glavas-etal-2014-hieve} contains 100 news articles. 
Within each article, annotations are given for both subevent membership and coreference relations. 
Using the same measurement of inter-annotator agreement (IAA) as event temporal relations in \citet{uzzaman-allen-2011-temporal}, the HiEve dataset has an IAA of 0.69 F1.

\stitle{Intelligence Community (IC)}
The IC corpus \cite{hovy-etal-2013-events} also contains 100 news articles annotated with membership relations. 
The articles 
report violence events such as attack, war, etc. 
We discard those relations involving implicit events annotated in IC, and calculate transitive closure for both subevent relations and co-reference to get annotations for all event pairs in text order as it is done for HiEve \cite{glavas-etal-2014-hieve}.

\stitle{Labeling \seg} \label{sec:obtain_seg}
We explain how to segment the document using annotations for subevent relations.
First, we use the annotated subevent relations (\textsc{Parent-Child} and \textsc{Child-Parent} only) to construct a directed acyclic event graph  for each document.
Due to the property of subevent relations, each connected component in the graph is actually a tree with one root node, which forms an event complex. 
If the graph constructed from document has one connected component, 
we remove the root node and separate the event graph into more than one event complexes. 
Since each event complex has a textual span in the document, we obtain several descriptive contexts that may or may not overlap with each other.
For those documents with non-overlapping descriptive contexts, 
their segmentations are therefore obtained.
In cases where two descriptive contexts of event complexes overlap with each other, if there exists such an event 
whose removal results in non-overlapping contexts, then we segment the contexts assuming this event is not considered. Otherwise, we merge the contexts into one segment.
Through this event-based text segmentation, on average we obtain 3.99 and 4.29 \textsc{EventSeg}s in the HiEve and IC corpus, respectively.

We summarize the data statistics in \Cref{tab:stats}.

\subsection{Baselines and Evaluation Protocols} \label{sec:baselines}
On IC dataset, we compare with two baseline approaches.
\citet{araki-etal-2014-detecting} propose a logistic regression model along with a voting algorithm for parent event detection.
\citet{wang-etal-2020-joint} use a data-driven model that incorporates handcrafted constraints with event temporal attributes to extract event-event relations.
On Hieve\footnote{Despite carefully following the details described in \citet{aldawsari-finlayson-2019-detecting} and communicating with the authors, we were not able to reproduce their results. Therefore, we choose to compare with other methods.}, we compare with a transformer-based language model \textsc{TacoLM} \cite{zhou-etal-2020-temporal} that fine-tunes on a temporal common sense corpora, and the method proposed by \citet{wang-etal-2020-joint} which also serves as the second baseline for IC.

We use the same evaluation metric on HiEve as previous methods \cite{zhou-etal-2020-temporal}, leaving 20\% of the documents out for testing\footnote{To make predictions on event complexes, we keep all negative \textsc{NoRel} instances in our experiments instead of strictly following \citet{zhou-etal-2020-temporal} and \citet{wang-etal-2020-joint} where negative  instances are down-sampled with a probability of 0.4.}. The $F_1$ scores of \textsc{Parent-Child} and \textsc{Child-Parent} and the micro-average of them are reported. 
In accordance with HiEve, the IC dataset is also evaluated with $F_1$ scores of membership relations instead of BLANC \cite{araki-etal-2014-detecting}, while the other settings remain the same with previous works.

\subsection{Experimental Setup} \label{sec:exp_setup}
We fine-tune the pre-trained 1024 dimensional RoBERTa \cite{liu2019roberta} to obtain contextual representations of event triggers in a supervised way given labels for membership relations and \seg.
Additionally, we employ 18 dimensional one-hot vectors for part-of-speech tags for tokens in documents to include explicit syntactic features in the model.
For each MLP we set the dimension to the average of the input and output neurons, following \citet{chen-etal-2018-yeji}.
The parameters of the model are optimized using AMSGrad \cite{reddi2018convergence}, with the learning rate set to $10^{-6}$.
The training process is limited to 40 epochs since it is sufficient for convergence.

\subsection{Results} \label{sec:results}
We report the results for subevent detection on two benchmark datasets, HiEve and IC, in \Cref{tab:results}.
Among the baseline methods, \citet{wang-etal-2020-joint} has the best results in terms of $F_1$ on both datasets. 
They integrate event temporal relation extraction, common sense knowledge and handcrafted logical constraints into their approach.
In contrast, our proposed method does not require constraints induced by domain experts, but still outperforms their $F_1$ score by 2.3 - 2.5\%.
We attribute this superiority to the use of connections between subevent relations and the linear discourse structure of segments. 
Thanks to the strong expressiveness of Rectifier Networks, we utilize these connections via the learning of linear constraints, thus incorporating incidental supervision signal from \seg.
Furthermore, the event pair representation in our model is obtained from broader contexts than the local sentence-level contexts for events in \citet{wang-etal-2020-joint}.
The new representation not only contains more information on events but naturally provides necessary clues for determining whether there is a break for \seg.

\begin{table}[]
\resizebox{\columnwidth}{!}{%
\begin{tabular}{l|l|ccc}
\hline
\toprule
                       &                           & \multicolumn{3}{c}{$F_1$ score} \\ 
Corpus                 & Model                     & PC       & CP       & Avg.    \\ \hline
\multirow{3}{*}{IC}    & \citet{araki-etal-2014-detecting}       & -        & -        & 0.262   \\
                       & \citet{wang-etal-2020-joint}        & 0.421    & 0.495    & 0.458   \\
                       & Our model                      & \textbf{0.446}    & \textbf{0.516}    & \textbf{0.481}   \\ \hline
\multirow{3}{*}{HiEve} 
 & \citet{zhou-etal-2020-temporal} & 0.485 &  0.494 & 0.489 \\
                       & \citet{wang-etal-2020-joint}       & 0.472    & \textbf{0.524}    & 0.497   \\
                       & Our model                      & \textbf{0.534}    & 0.510    & \textbf{0.522}   \\ 
                       \bottomrule
\end{tabular}
}
\caption{Experimental results for subevent detection on IC and HiEve corpus. \textsc{PC}, \textsc{CP} and Avg. denote \textsc{Parent-Child}, \textsc{Child-Parent} and their micro-average, respectively. $F_1$ scores for \textsc{PC} and \textsc{CP} are not reported in \citet{araki-etal-2014-detecting}. }
\label{tab:results}
\end{table}


We further perform an ablation analysis to aid the understanding of the model components and report our findings in \Cref{tab:Ablation}. 
Without any constraints, integrating \seg prediction as an auxiliary task brings along an absolute gain of 0.2\% and 0.6\% in $F_1$ on HiEve and IC respectively over the vanilla single-task model with RoBERTa fine-tuning.
This indicates that \seg information is beneficial to the extraction of membership relations.
When membership constraints are added via the regularization term into the loss function, the model's performance on subevent detection is significantly improved by 2.1\% in $F_1$ on HiEve dataset.
Incorporating constraints involving two tasks further enhances the model performance by 0.5\% - 1.1\%.
This indicates that the global consistency ensured within and across \textsc{EventSeg}s is important for enhancing the comprehension for subevent memberships.

\begin{table*}[!t]
    \centering
    {
    \small
    \begin{tabular}{l|ccc|ccc}\hline 
    \toprule
    & \multicolumn{3}{c|}{HiEve} &  \multicolumn{3}{c}{IC} \\
    Model & $P$ & $R$ & $F_1$ & $P$ & $R$ & $F_1$ \\ \hline
   Single-task Training &  43.9 & \textbf{56.6} & 49.4 & 44.5 & 46.9 & 45.8 \\
   
   Joint Training & 45.7 & 54.2 & 49.6 & 39.9 & 56.5 & 46.4 \\
    \hline
    + Membership Constraints & \textbf{55.6} & 48.5 & 51.7 &  \textbf{50.1} & 45.8 & 47.0 \\
    + Membership   + \seg & 51.9 & 53.6 & \textbf{52.2} & 39.6 & \textbf{64.0} & \textbf{48.1} \\  

    \bottomrule
    \end{tabular}
    }
    \caption{Ablation study results for subevent detection. The results on both datasets are the micro-average of \textsc{Parent-Child} and \textsc{Child-Parent} in terms of precision, recall, and $F_1$.
    ``+ Membership Constraints'' denotes adding automatically learned constraints for membership relations upon the joint training model.
    The row of ``+ Membership   + \seg'' shows the results of the complete model.
    }
    \label{tab:Ablation}
\end{table*}

\subsection{Case Study for \seg Prediction}
\label{sec:case_study}
Here we provide an analysis of model performance on the task of \seg prediction.
Though \seg prediction is somewhat different from text segmentation in concept, we can use methods for text segmentation as baselines for \seg prediction. 
We train a recent BERT-based model \cite{lukasik-etal-2020-text} for text segmentation based on annotations for \seg in the HiEve and IC corpora and compare our method with this baseline.
In \Cref{tab:seg_prediction} we show the performances of the baseline model and ours for \seg prediction in terms of $F_1$ on HiEve and IC.
Since our solution for \seg prediction is essentially similar to the cross-segment BERT model in terms of representations of segments, our performance is on par with the baseline model.

\begin{table}[!t]
\centering
\setlength\columnsep{1pt}
\resizebox{\columnwidth}{!}{%
\begin{tabular}{l|c|c}
\hline
\toprule
          Model                            & HiEve  & IC     \\ \hline
Cross-segment BERT \cite{lukasik-etal-2020-text} & 55.2 & 58.3\\
Our model & 56.8 & 57.4 \\ 
\bottomrule
\end{tabular}
}
\caption{\seg prediction performance in terms of $F_1$ on the HiEve and IC corpus.}
\label{tab:seg_prediction}
\end{table}

\subsection{Analysis on Constraint Learning} \label{sec:exp_cons}

We further provide an in-depth qualitative analysis on different types of logical constraints captured by the constraint learning.

\subsubsection{Types of Learned Constraints}\label{sec:constraint_type}
We expect that both task-specific constraints (membership relations only) in previous works \cite{glavas-snajder-2014-constructing, wang-etal-2020-joint} and cross-task constraints can be automatically captured in our framework. Accordingly, we separately analyze these two constraints.

\stitle{Task-specific Constraints.} 
Since we are using three-event subgraph for constraint learning, apparently, transitivity constraints for membership relations like
\begin{equation*}
    \begin{split}
          Y_{i, j}^r   +  Y_{j, k}^r & -  Y_{i, k}^r \le 1, \\
          r \in \{\textsc{Parent-Child}, &  \textsc{Child-Parent}, \textsc{Coref}\},
    \end{split}
    \label{eq:trans}
\end{equation*}
can be learned; whereas constraints that typically involve two events, e.g., symmetry constraints for membership relations like 
\begin{equation*}
    \begin{split}
        Y_{i, j}^r  = & Y_{j, i}^{\bar{r}}, \\
        r \in \{\textsc{Parent-Child} & ,\textsc{Child-Parent}\},
    \end{split}
    \label{eq:sym}
\end{equation*}
can also be learned by assigning the third event $e_k$ to the same event as $e_i$ and treating the relation of $(e_i, e_k)$ as \textsc{Coref}.

\stitle{Cross-task Constraints.}
Here we provide an analysis of cross-task constraints for both membership relations and \seg information learned in the model.
We give an example constraint in the form of linear inequality learned from HiEve 
\begin{align*}
    \begin{split}
         &0.13x_0 +0.19x_1 +0.27x_2 +0.08x_3 -0.18x_4 \\
         +&0.09x_5 +0.13x_6 +0.25x_7 +0.04x_8 -0.18x_9 \\
        +& \cdots + 0.02x_{18} +0.07x_{19} + \cdots + 0.05
        \ge 0,
    \end{split}
\end{align*}
where $x_1$ and $x_6$ denote the variables for $Y_{i, j}^r = 1$ and $Y_{j, k}^r = 1$ ($r = $ \textsc{Child-Parent}) respectively, and they both have positive coefficients.
If we look at expected labels for $\mathcal{P}(\mathbf{X}_{i, k})$, we can see that $x_{18}$ and $x_{19}$ which denote the variables for $Y_{i, k}^r = 1, Z_{i, k} = 0$ and $Y_{i, k}^r = 1, Z_{i, k} = 1$ have coefficients of 0.02 and 0.07, respectively. 
The two positive coefficients for $x_{18}$ and $x_{19}$ indicate that \textbf{(a)} $(e_i, e_k)$ is possible to have \textsc{Child-Parent} relation, and \textbf{(b)} the possibility of $(e_i, e_k)$ being in the same \seg is greater than two events being in different \textsc{EventSeg}s.

\subsubsection{Qualitative Analysis}

We set $K$ to 10 since we observe less number of constraints will decrease the performance of learning accuracy while increasing $K$ 
does not cause noticeable influence.
We optimize the parameters using Adam with a learning rate of 0.001 and the training process is limited to 1,000 epochs.
We show the performance of constraint learning in \Cref{tab:cons_learn}.
Since the constraints for membership relations should be declarative hard constraints like symmetry and transitivity constraints in \Cref{sec:constraint_type}, the accuracy of constraint learning is equal or close to 100\%.
Yet, those hard-to-articulate constraints that incorporate \seg information are more difficult to learn, and thus the Rectifier Network has a less satisfying performance in terms of accuracy on the test set of HiEve and IC (96.44\% and 98.01\%).

\begin{table}[!t]
\centering
{%
\small
\begin{tabular}{l|c|c}
\hline
\toprule
          Constraints                            & HiEve  & IC     \\ \hline
Membership                      & 99.13 & 100.00       \\ 
Membership   + \seg & 96.44 & 98.01 \\ 
\bottomrule
\end{tabular}
}
\caption{Constraint learning performance in terms of accuracy on test set. ``Membership'' denotes the constraints involving membership relations only, while ``Membership + \seg'' denotes full constraints.}
\label{tab:cons_learn}
\end{table}

\section{Conclusion}

In this work we propose an automatic and efficient way of learning and enforcing constraints for subevent detection.
By noticing the connections between subevent dection and \seg, we adopt \seg prediction as an auxiliary task which provides effective incidental supervision signals.
Through learning and enforcing constraints that can express hard-to-articulate constraints, the logical rules for both tasks are captured to regularize the model towards consistent inference.
The proposed approach outperforms SOTA data-driven methods on benchmark datasets and provides comparable results with recent text segmentation methods on \seg prediction.
This demonstrates the effectiveness of the framework on subevent detection and the potential of solving other structured predictions tasks in NLP.

\section*{Ethical Considerations}
This work does not present any direct societal consequence. The proposed method aims at supporting high-quality extraction of event complexes from documents with the awareness of discourse structures and automated constraint learning. We believe this study leads to intellectual merits of developing robust event-centric information extraction technologies. It also has broad impacts, since constraints and dependencies can be broadly investigated for label structures in various natural language classification tasks. The acquired eventually knowledge, on the other hand, can potentially benefit various downstream NLU and NLG tasks.

For any information extraction methods, real-world open source articles to extract information from may include societal biases. 
Extracting event complexes from articles with such biases may potentially propagate the bias into acquired knowledge representation.
While not specifically addressed in this work, the ability to incorporate logical constraints and discourse consistency can be a way to mitigate societal biases.

\section*{Acknowledgement}

We appreciate the anonymous reviewers for their insightful comments.

This research is supported by the Oﬃce of the Director of National Intelligence (ODNI), Intelligence Advanced Research Projects Activity (IARPA), via IARPA Contract No. 2019-19051600006 under the BETTER Program, and by contract FA8750-19-2-1004 with the US Defense Advanced Research Projects Agency (DARPA). The views expressed are those of the authors and do not reflect the official policy or position of the Department of Defense or the U.S. Government.

\bibliography{anthology,custom,ccg}
\bibliographystyle{acl_natbib}

\end{document}